\documentclass{article}

\ifdefined\pdfobjcompresslevel
\fi

\usepackage[preprint]{neurips_2026}

\usepackage[utf8]{inputenc} 
\usepackage[T1]{fontenc}    
\usepackage{hyperref}       
\usepackage{url}            
\usepackage{booktabs}       
\usepackage{amsfonts}       
\usepackage{nicefrac}       
\usepackage{microtype}      
\usepackage{xcolor}         

\usepackage{graphicx}
\usepackage{wrapfig}

\usepackage{booktabs}
\usepackage{multirow}
\usepackage{array}
\usepackage{caption}
\usepackage{subcaption}
\usepackage{enumitem}
\usepackage{amsmath}
\usepackage{todonotes}
\usepackage{amssymb}
\usepackage{natbib}
\usepackage{cleveref}

\usepackage[normalem]{ulem} 
\usepackage{xcolor}         
\usepackage{bbm}

\usepackage{algorithm}
\usepackage{algorithmicx}

\newcolumntype{x}[1]{>{\centering\let\newline\\\arraybackslash\hspace{0pt}}p{#1}}

\title{Information theoretic underpinning of self-supervised learning by clustering}

\author{Josef Kittler$^1~~~~$ \qquad Sara Atito$^{1,2}$ \qquad $~~~~$Muhammad Awais$^{1,2}$\\
$^1$Centre for Vision, Speech and Signal Processing (CVSSP),\\
$^2$Surrey Institute for People-Centred AI (PAI), \\
University of Surrey, Guildford, Surrey GU2 7XH, UK\\
{j.kittler,sara.atito,muhammad.awais}@surrey.ac.uk
}

\begin{document}

\maketitle

\begin{abstract}
 Self-supervised learning (SSL) is recognized as an essential tool for building foundation models for Artificial Intelligence applications. The advances in SSL have been made thanks to vigorous arguments about the principles of SSL and through extensive empirical research. The aim of this paper is to contribute to the development of the underpinning theory of SSL, focusing on the deep clustering approach. By analogy to supervised learning, we formulate SSL as K-L divergence optimization. 
 The mode collapse is prevented by imposing an optimisation constraint on the teacher distribution. This leads to  normalization using inverse cluster priors.  We show that using Jensen inequality this normalization simplifies to the popular batch centering procedure.  Distillation 
 and centering are common {heuristics-based} practices in SSL, {but our work underpins them theoretically.}  The theoretical model developed not only supports specific existing successful SSL methods, but also suggests directions for future investigations. 
\end{abstract}

\section{Introduction}

Learning without labels is a problem that goes back to AI prehistory. Its origins can be traced back to Karl \cite{Pearson}, who was interested in identifying the structure of observed data by fitting to it a Gaussian mixture model. The early machine learning solutions date back to the k-means algorithm \cite{Forgy,MacQueen} and its soft version in the form of the EM \cite{Rubin} algorithm for Gaussian mixture modelling. The evolution of unsupervised learning over the fifty years since the k-means algorithm is documented in Jain’s historical overview \cite{Jain-prl2010}. The most recent developments in this unsupervised learning area tackle the challenge of multi-view clustering, and the complexity of its view unaligned formulation \cite{stegmuller2023croc
}.

The advent of deep learning transformed the problem of learning without labels to self-supervised learning. The motivation for self-supervised learning goes well beyond the discovery of underlying data structure, which is the primary objective of unsupervised learning. It has been inspired by the limitations of supervised learning, which, first of all, requires expensive training data annotation, and second, and most importantly, it focuses on class discrimination, rather than on capturing and characterising the data properties. For these reasons supervised learning is not an ideal basis for building foundation models. In contrast, unlabelled data required by self-supervised learning is relatively easy to collect. Crucially, the nature of self-supervised learning, thanks to its pre-text tasks, learns about the data properties, such as similarity of neighbouring pixels,  image structure, context, and environmental variations. It is therefore better suited for learning good representation \cite{fukushima1980neocognitron,wiskott2002sfa,hinton2006dbn}, which is more transferable to other domains and better equipped for building foundation models.

Many different learning principles have been suggested for self-supervised learning. These include the idea of autoencoding, which aims to learn concise deep representation by minimising the reconstruction error \cite{Vincent2010}. The reconstruction task can be made more challenging by masking some of the input data to force more vigorous learning with the aim of improving performance \cite{atito2021sit,
he2022masked,xie2022simmim}. Most methods learn by creating alternative views of the input and seek representation that is invariant to view changes. This can be achieved simply by maximising the invariance of the representation to distinct views,  as in BYOL \cite{grill2020bootstrap}, or by contrasting the invariance of distinct views with their dissimilarity from other instances of the training data \cite{oord2018cpc}, or their proxies \cite{caron2018deep}. The Barlow Twins \cite{zbontar2021barlow} is based on the principle of minimum redundancy, while \cite{Tishby1999} advocates the concept of information bottleneck measured in terms of mutual information between the input and output probability distributions. 

Many SSL methods initially emerged as heuristic procedures, using loss functions tailored to promote their adopted learning principle. Their algorithmic success has been followed with diverse attempts to provide a theoretical insight and underpinning of their inherent learning process. Mutual information figures prominently as the underlying mathematical model \cite{oord2018cpc,saunshi2019theory,Asano-iclr2020,Chen2020}, seeking neural network parameters producing the same latent representation for a training sample and its augmented versions. Another starting point is information bottleneck \cite{Li2025}.  \cite{Sansone2025} categorised SSL methods into contrastive learning and clustering approaches. They showed that the joint probability likelihood of original and augmented observations can be lower bounded by a loss function used for clustering-based SSL.  \cite{Tan2024}  groups SSL methods into contrastive learning, features decorrelation methods and masked image modelling, and uses matrix information theory to establish their relationship..  

The common feature of many SSL approaches is an interaction of the on-line and target branches of the siemese network accomplishing the learning. This is similar to a student-teacher interaction in distillation learning, yet none of the existing SSL methods formulate the learning problem directly to reflect this relationship.  

In this paper, our starting point is the formulation of SSL as distillation learning, where the student network learns from teacher in a conventional way by minimising the K-L divergence, rather than just cross-entropy.  In contrast with supervised learning, where the teacher distribution is fixed, in SSL we have to learn both the student, as well as the teacher distribution.  This will be achieved by alternating direction optimisation. This formulation is closely linked to a specific family of SSL by clustering \cite{caron2018deep,caron2020swav,caron2021emerging,oquab2023dinov2,simeoni2025dinov3,zhou2021ibot}, but it has implications for other principles of self-supervised learning as well. To avoid the mode collapse, the solution will be regularised, as in \cite{caron2020swav}. Rather than calling on an algorithmic solution to the regularisation problem, as in \cite{caron2020swav}, we develop the regularised K-L divergence formulation mathematically and derive a close form solution to the teacher network normalisation problem. It will be shown that this leads to a normalisation method, which involves scaling by the cluster priors. This form of normalisation was used in  \cite{Amrani2022}, but in their case justified on purely heuristic basis. It has been shown in \cite{Amrani2022} that this normalisation achieves learning without a mode collapse. In this paper, the significance of our derivation is that it enables us to establish a link with the mode-collapse inhibiting procedure by means of centering \cite{caron2021emerging}. This provides a theoretical justification for the popular and computationally effective normalisation procedure in SSL.

In summary, we formulate SSL as distillation learning using K-L divergence as a learning objective. The learning process estimates the cluster weights and the network parameters to optimise the student distribution, and defines the update process for learning the teacher.  The mathematical model developed provides a theoretical underpinning for both, the clustering approach to SSL, and the heuristic normalisation method based on data centering.

\section{Related Work}\label{sec:related}
The common distingushing feature of SSL methods is that the unsupervised learning task is converted into  supervised learning by means of pre-text tasks. Generally speaking, the input data is modified by a selected transformation and the deep network is trained to respond to the modified input activation in the same way as to the original data.  Examples include colourisation \cite{zhang2016colorful,larsson2016learning,larsson2017colorization}, relative patch location \cite{doersch2015unsupervised}, solving jigsaw puzzles \cite{noroozi2016unsupervised,kim2018learning}, cross-channel prediction \cite{zhang2017split}, predicting noise \cite{bojanowski2017unsupervised}, predicting image rotations \cite{gidaris2018unsupervised}, spotting  artefacts \cite{jenni2018self}, etc. 
These early efforts were later strengthened by the emergence of 
instance discrimination methods, where the similarity of augmented views of one instance in a batch is contrasted with that of other instances in the batch \cite{wu2018unsupervised,Chen2020}. 

In parallel, clustering-based SSL offered another direction. DeepCluster \cite{caron2018deep} and SwAV \cite{caron2020swav} performed online clustering of features and used cluster assignments as pseudo-labels. 
Later, the clustering idea has been pursued in studies involving the  Vision Transformers (ViTs) architecture \cite{caron2021emerging,zhou2021ibot}. DINOv2 \cite{oquab2023dinov2} demonstrated the scalability of this approach to a huge training set, comprising hundreds of millions of curated images. Similarly, the scalability to multi-billion parameter models with training refinements like Gram anchoring was demonstrated by DINOv3 \cite{simeoni2025dinov3}. 
 
The necessity of contrastive learning was questioned by later work, which showed that negative pairs were not strictly necessary. In BYOL's framework \cite{grill2020bootstrap} the learning of effective features is accomplished using only positive pairs by relying on batch normalisation \cite{ioffe2015batch}, while SimSiam \cite{chen2021exploring} simplified the setup further by relying on weight sharing and a stop-gradient operation to prevent collapse. Another non-contrastive objective is adopted by Barlow Twins \cite{zbontar2021barlow}, which draws on the concept of redundancy of representation, while VICReg \cite{bardes2022vicregl} focuses on variance–covariance regularisation.  

Within the above frameworks, the recent refinements focused on masked image modelling \cite{atito2021sit,
he2022masked,xie2022simmim} and sharpening of image content representation by developing dense-focused methods
\cite{xie2021unsupervised,henaff2021efficient,stegmuller2023croc,lebailly2024cribo,ukic2025objectcentricOCEBO}, providing better concept localisation.

In this paper we do not propose a new learning model. Rather, the aim is to underpin existing approaches to self-supervised learning by a comprehensive information theoretic model. We focus on the clustering approach in particular, but the development has a wider implication for SSL. 


\section{Preliminaries}
\label{prelim}

An unlabelled dataset, collected by sensing the universe surrounding us, represents multiple concepts. The data relating to one concept is expected to cluster together and, at the same time, to be very distinct from that of other concepts. The aim of clustering is to discover this structure in the data without the help of semantic labels. 
In the case of deep learning, the clustering problem can be formulated in a representation space, or some projection of it. In this paper we are not concerned with the merits of one or the other. Whatever the choice of space is, we assume that an input data point $\mathbf{x}$ is mapped to its corresponding vector $\mathbf{z}$ by our deep network parametrised by network parameters $\theta$. We assume that within this space the concepts captured by the data are represented by $K$ clusters, each modelled by a Gaussian with mean $\mathbf{w}_i,i=1,...,K$ and isotropic variance (temperature) $\tau$. In other words, the task is to fit a mixture of Gaussians $ p(\mathbf{z})=\frac{1}{\beta}\sum_{y=1}^K\exp{\{\frac{-1}{\tau}(\mathbf{z}-\mathbf{w}_y)^T(\mathbf{z}-\mathbf{w}_y)\}} P(y)$
to our training data in this space, where $P(y)$ is the prior probability mass of data in cluster $y$ and $\beta$ is a scaling parameter to ensure that $p(\mathbf{z})$ satisfies axiomatic properties of probability density.  

Clustering is basically a classification task, where both the training data assignment to clusters and cluster models are learnt during training. A solution to the task is typically realised by creating two deep networks: one performing the role of a teacher, and the other, the role of a student learning from the teacher. The teacher assigns a training sample $\mathbf{z}_i$ to cluster $y$ with probability $Q(y|\mathbf{z}_i)$ and the student strives to mimic this by aligning its probability of assignment $P(y|\mathbf{z}_i)$ to  $Q(y|\mathbf{z}_i)$ by adjusting its network parameters $\theta$ and the cluster means $\mathbf{w}_y, \forall y$. Under our Gaussian assumption the student posterior $P(y|\mathbf{z})$ is given by
\begin{equation}
    \label{it7a}
    P(y|\mathbf{z})= \frac{\exp{\{\frac{-1}{\tau}(\mathbf{z}-\mathbf{w}_y)^T(\mathbf{z}-\mathbf{w}_y)\}} P(y)} {\sum_{j=1}^K \exp{\{\frac{-1}{\tau}(\mathbf{z}-\mathbf{w}_j)^T(\mathbf{z}-\mathbf{w}_j)\}} P(j)}
\end{equation}
In supervised learning the teacher distribution is a one-hot vector. In our case, $Q(y|\mathbf{z})$ is a soft distribution. 

The common objective for distillation learning is 
the Kullback-Leibler divergence
\begin{equation}
\label{it3}
\begin{array}{l}
J_{KL}(Q||P)=\sum_{i=1}^N \sum_{y=1}^K Q(y|\mathbf{z}_i)  \log \frac{Q(y|\mathbf{z}_i)}{P(y|\mathbf{z}_i)}
=\sum_{i=1}^N \sum_{y=1}^K Q(y,
|\mathbf{z}_i) \log Q(y|\mathbf{z}_i)- \\
-\sum_{i=1}^N \sum_{y=1}^K Q(y|\mathbf{z}_i) \log{P(y|\mathbf{z}_i)}
\end{array}
\end{equation}
where $N$ is the size of the dataset.
Clearly, the criterion function will be minimised if the argument of the $\log$ function is one. This will be satisfied whenever $Q(y|\mathbf{z}) = P(y|\mathbf{z})$.  


In clustering the supervisory signal is unknown, and has to be estimated. We shall optimise (\ref{it3}) by the alternating direction method. Given $Q(y|\mathbf{z})$, we optimise (\ref{it3}) with respect to $\theta$ and $\mathbf{w}_i, \forall i$. As 
$Q(y|\mathbf{z})$ is fixed, this requires minimisation of the cross entropy term, i.e. at point $\mathbf{z}$
\begin{equation}
\label{it4a}
    \nabla_{\mathbf{w}_y}J = -\frac{Q(y|\mathbf{z})}{P(y|\mathbf{z})}\nabla_{\mathbf{w}_y} P(y|\mathbf{z})
\end{equation}
Referring to eq. (\ref{it7a})
\begin{equation}
\label{it4b}
    \nabla_{\mathbf{w}_y} P(y|\mathbf{z}) = -\frac{1}{\tau}
    P(y|\mathbf{z})[1-P(y|\mathbf{z})](\mathbf{z}-\mathbf{w}_y)
\end{equation}
Thus, the gradient of $J_{KL}(Q||P)$ with respect $\mathbf{w}_y$  
\begin{equation}
\label{it4}
    \nabla_{\mathbf{w}_y}J \propto \frac{1}{\tau}Q(y|\mathbf{z})[1-P(y|\mathbf{z})](\mathbf{z}-\mathbf{w}_y)
\end{equation}
and similarly  with respect to $\theta$,

From (\ref{it4}) it is evident the direction of the gradient would be dominated by cluster mean vectors of large magnitude. This suggests the length of the mean vectors should be normalised. In fact it is a common practice to set all mean vectors to a unit length. Similarly, even if the mean vectors are normalised, if the magnitude of instance vector $\mathbf{z}$ is unchecked, (\ref{it4}) suggests it could dominate the cluster mean updating process. To avoid any learning difficulties, it is important that the deep representation data is routinely normalised, typically to having isotropic unit standard deviation. In fact, frequently the data and the class means are normalised to lie on a hypersphere of radius one, which takes care of the requisite normalisation. Under this normalisation the cluster models simplify and the cluster assignment probability distributions in (\ref{it7a}) become softmax functions where the priors are ignored, i.e.
\begin{equation}
    \label{it8}
    P(y|\mathbf{z})= \frac{\exp{\{\frac{1}{\tau}(\mathbf{z}^T\mathbf{w}_y)\}}}{\sum_{j=1}^K \exp{\{\frac{1}{\tau}(\mathbf{z}^T\mathbf{w}_j)}\}}
\end{equation}
The cluster mean vectors in (\ref{it8}) are commonly referred as the cluster weight vectors. We shall confine our ensuing discussion to this definition of cluster assignment posterior $P(y|\mathbf{z})$. 

Once the network parameters and cluster means are updated, we can fix $P(y|\mathbf{z}), \forall y$ and change the direction of optimisation to focus on $Q(y|\mathbf{z}), \forall y$. Differentiating our K-L divergence in (\ref{it3}) with respect to $Q(y|\mathbf{z})$ and setting the derivative to zero leads to
\begin{equation}
    \label{it7}
    Q(y|\mathbf{z})= P(y|\mathbf{z})
\end{equation}
In fact, this solution is not surprising as it is well known that K-L divergence \cite{kullback1997information} becomes zero when the two probability distributions are identical.
In practice, eq. (\ref{it7}) implies that, after a certain number of iterations, the teacher network adopts the parameters $\theta$ and $\mathbf{w}_j$ from the student. This can be done by direct copying or, for the sake of stability, by a momentum update.   

The common practice in SSL is that the weight vectors are normalised to a unit norm, the data distribution is scaled to a unit standard deviation, the temperature of the teacher is higher than that of a student and for every training instance $\mathbf{x}$ the student is fed its augmented version $\tilde{\mathbf{x}}$. By alternating direction optimisation the learning process will iteratively improve the objective function. However, the learning could lead to a mode collapse. This tendency can be prevented by regularisation. This will be the topic of discussion in the next section.

\section{Learning the supervision
signal using regularised K-L divergence}
\label{rem}
The most obvious way to prevent mode collapse is to force the SSL algorithm to distribute the training data, projected into the embedding space by the DNN, uniformly over the whole latent space. A vehicle for this distribution task are the clusters. In other words, we wish to distribute the assignment of the data to clusters by the supervisory signal in an equitable way. This is not only an argument for preventing a mode collapse, but an intuitively reasonable proposition, that distinct concepts in the training data should cover the complete embedding space, rather than just a part of it. 

Our aim is to learn a representation that will distribute the training data uniformly over a specified number of cluster. Mathematically, this means that the prior probabilities $Q(y)$ for the training data assigned to different clusters $y=1,...,K$ are roughly the same, i.e.
\begin{equation}
    \label{it11}
    Q(y)= \frac{1}{N}\sum_{j=1}^NQ(y|\mathbf{z}_j) \approx \frac{1}{K}
\end{equation}
This objective can be formulated as a regularisation constraint in the form of entropy
$-\sum_{y=1}^K Q(y)\log Q(y)$, which becomes maximum when the prior probabilities are uniformly distributed. This can be encouraged by regularising the clustering solution so that entropy of the means of the cluster assignments is maximised. This follows the principle advocated in \cite{caron2020swav}, but rather than resorting to a pre-canned algorithm such as the Sinkhorn-Knopp optimal transport optimisation, we develop the formulation mathematically. 
 
Since our self-supervised learning  is formulated as a minimisation problem, the entropy constraint, $-\sum_{y=1}^K Q(y)\log{Q(y)} $ has to be inserted with the minus sign, yielding an overall objective function as
\begin{equation}
    \label{rem1}
    \begin{array}{l}
\mathcal{L}(Q||P)=
\sum_{j=1}^N\sum_{y=1}^K Q(y|\mathbf{z}_j) \log Q(y|\mathbf{z}_j) 
-\sum_{j=1}^N\sum_{y=1}^K Q(y|\mathbf{z}_j)\log P(y|\mathbf{z}_j) \\
+\sum_{y=1}^K \sum_{j=1}^N Q(y|\mathbf{z}_j)\log{\sum_{j=1}^N Q(y|\mathbf{z}_j)} \\
\end{array}
\end{equation}
In addition to the negative entropy constraint, we also have to ensure that the supervisory signal satisfies the axiomatic properties of probabilities, namely that $\sum_{y=1}^K Q(y|\mathbf{z})=1$. This can be imposed by a loss function $\frac{1}{2}[\sum_{y=1}^K Q(y|\mathbf{z})-1]^2$. Thus our final objective function will be given as  
\begin{equation}
    \label{rem11}
    \begin{array}{l}
\mathcal{L}(Q||P)=
\sum_{j=1}^N \sum_{y=1}^K Q(y|\mathbf{z}_j) \log Q(y|\mathbf{z}_j) 
-\sum_{j=1}^N\sum_{y=1}^K Q(y|\mathbf{z}_j)\log{P(y|\mathbf{z}_j)} \\
+\sum_{y=1}^K \sum_{j=1}^N Q(y|\mathbf{z}_j)\log{\sum_{j=1}^N Q(y|\mathbf{z}_j)}  + \frac{1}{2}\sum_{j=1}^N[\sum_{y=1}^K Q(y|\mathbf{z}_j)-1]^2\\
\end{array}
\end{equation}

In the regularised K-L divergence we need to optimise the predictor probablity with respect to the deep neural network parameters $\theta$ and $\mathbf{w}_i$. The only term that is a function of $P(y|\mathbf{z})$ is
\begin{equation}
    \label{rem21}
L(\theta, \mathbf{w}_i)=
-\sum_{j=1}^N\sum_{y=1}^K Q(y|\mathbf{z}_j)\log{P(y|\mathbf{z}_j)}
\end{equation}
which is a standard crossentropy loss function. However, the supervisory signal $Q(y|\mathbf{z})$ is also unknown. To optimise the K-L divergence in (\ref{rem11}), as in the previous section, we adopt the alternating direction optimisation method, where we first fix $Q(y|\mathbf{z})$, and optimise $P(y|\mathbf{z})$ using the loss function in (\ref{rem21}) for several iterations. Then we fix $P(y|\mathbf{z})$, i.e. fix parameters $\theta$ and $\mathbf{w}_i$, and optimise $Q(y|\mathbf{z})$. This process is iterated until convergence. 

In particular, we optimise the teacher network based on the gradient of our objective function.   
Taking the derivative of the K-L divergence in (\ref{rem11}) at point $\mathbf{z}$  with respect to $Q(y|\mathbf{z})$
gives 
\begin{equation}
\label{it6}
\begin{array}{l} 
\frac{d}{dQ} \mathcal{L}(Q||P)= \log Q(y|\mathbf{z}) -\log P(y|\mathbf{z}) 
 +\log\frac{1}{N}\sum_{j=1}^N Q(y|\mathbf{z}_j) 
+ \sum_{y=1}^K \frac{1}{N}\sum_{j=1}^N Q(y|\mathbf{z}_j) +1
\end{array}
\end{equation}
As the aim is to minimise divergence, the derivative with respect to $Q$ should be zero, which will be achieved if 
\begin{equation}
    \label{it70}
  Q(y|\mathbf{z})\frac{1}{N}\sum_{j=1}^NQ(y|\mathbf{z}_j) = \frac{P(y|\mathbf{z})}{\exp\{\sum_{y=1}^K \frac{1}{N} \sum_{j=1}^N Q(y|\mathbf{z}_j) +1\}}
\end{equation}
Note that the exponent in the denominator on the right hand side of eq. (\ref{it70}) is the same for all cluster identities, and can, therefore, be ignored. Further, the second term on the left hand side is independent of $\mathbf{z}$. It actually measures the a priori probability $Q(y)$ of samples belonging to cluster $y$. Thus rearranging, the supervisory signal (the teacher) follows the predictor probabilities (student), subject to a scaling factor, i.e. 
\begin{equation}
    \label{it73}
    Q(y|\mathbf{z}) = \frac{P(y|\mathbf{z})}{Q(y)}
\end{equation}

When optimising the predictor probability in (\ref{rem21}), we use the teacher network parameters $\theta_T,\mathbf{W}_T$ where $\mathbf{W}=(\mathbf{w}_1,...,\mathbf{w}_K)$, for the supervision of the student network to update the network parameters $\theta_S,\mathbf{W}_S$.  After $n_e$ epochs, these network parameters are passed on to the teacher, potentially using the exponential moving average procedure. 
Using the new teacher parameters, the prior probability $Q(y)$ is estimated as
\begin{equation}
    \label{it9a}
  Q(y) =  \frac{1}{N}\sum_{j=1}^N P(y|\mathbf{z})_{|\theta_T,W_T}
\end{equation}
and 
\begin{equation}
   \label{it9b}
  Q(y|\mathbf{z}) =\frac{P(y|\mathbf{z})_{|\theta_T,W_T}}{Q(y)}
\end{equation}

As intended, this updating process will encourage the training data to be distributed evenly over all the clusters. Empty clusters will have a low prior $Q(y)$, and their posterior will be boosted by $Q(y)^{-1}$, while highly populated clusters will be unaffected. However, as the teacher probabilities $Q(y|\mathbf{z}), y=1,...k$ have to sum up to one, they have to be normalised, and this will promote sample reassignment to the clusters with lower priors. 

Our information theoretic approach resulted in a supervised learning procedure, where the teacher supervisory signal is updated during learning. Basically, the learning process realises distillation, where both student and teacher are updated iteratively. The teacher updates its supervisory signal based on the current student network parameters, but weighted so as to promote equitable distribution of data over clusters. The updated teacher probabilities drive the update of the student neural network parameters. The network-update aims to align the student probabilities to the teacher output. 

The cluster weight vector update could be carried out simultaneously with the network learning in a batch mode. However, batches contain a relatively small number of samples. The drawback of the small sample problem is that  the estimates of the cluster priors will be very noisy, and even more seriously, not all clusters will have representative samples within a batch. This would make it difficult to differentiate between genuinely void clusters as a result of a limited number of concepts, and the clusters that are void by virtue of data paucity. Our preferred approach is to update the cluster priors using all training data. The weight vectors could be updated at the same time  using standard clustering techniques, such as the EM or k-means algorithm. However, this weight update option will be considered in a future investigation, as this may have to be done more frequently, than the cluster prior computation for the supervision signal update. For the current discussion we assume that the weight vector update is done by gradient descent. 



To recapitulate, our distillation approach requires the teacher network 
to retain its network parameters and the updated cluster weights for the next $n_e$ epochs of the student learning. The value of $n_e$ is determined by two factors: In the first instance it is the computational complexity, as the update of the cluster weight vectors requires an update involving many training samples, in order to have a realistic estimate of the cluster priors. The other consideration is the noisiness of the evolution of the network updates as a function of the number of epochs. To minimise the impact of these  variations, either the number of epochs $n_e$ between the teacher updates should be sufficient to smooth them out, or the value of $n_e$ could be relatively small, provided we rely on a momentum update for the teacher network parameters. 







\section{Relationship to Dino}

The aim of this section is to look at Dino clustering method and try to establish its relationship with the proposed information theoretic approach.  We start by revisiting an early version of deep clustering, which normalised the teacher by {\bf centering}.

 Let us return to (\ref{it8}) and substitute $P(y|\mathbf{z})$ by its softmax definition, i.e. 
\begin{equation}
    \label{it15}
    Q(y|\mathbf{z}) = \frac{\frac{\exp{\{\mathbf{z}^T\mathbf{w}_y/\tau_T}\}}{\sum_i^K \exp{\{\mathbf{z}^T\mathbf{w}_i/\tau_T}\}}}{\frac{1}{N}\sum_{j=1}^N \frac{\exp{\{\mathbf{z}_j^T\mathbf{w}_y/\tau_T}\}}{\sum_i^K \exp{\{\mathbf{z}_j^T\mathbf{w}_i/\tau_T}\}}}
\end{equation}
We can analyse it more closely by considering the first normalising term $\sum_i^K \exp{\{\mathbf{z}^T\mathbf{w}_i/\tau_T\}}$. The exponential function is convex, and hence using Jensen's inequality, we can write
\begin{equation}
    \label{it16}
    \begin{array}{l}
\sum_i^K \exp{\{\mathbf{z}^Tw_i/\tau_T\}}  \ge  K \exp{\{\mathbf{z}^T\frac{1}{K}\sum_{i=1}^K \mathbf{w}_i/\tau_T\}}
=\exp{\{K \mathbf{z}^T\bar{\mathbf{w}}/\tau_T\}}
\end{array}
\end{equation}
where $\bar{\mathbf{w}}$ is the mean of the cluster weight vectors, i.e.
\begin{equation}
\label{it16a}
\bar{\mathbf{w}}= \frac{1}{K} \sum_{i=1}^K \mathbf{w}_i
\end{equation}
If we ensure that the weight vectors are centred, i.e. set $\bar{\mathbf{w}}=0$, then the normalising term can be approximated by  
\begin{equation}
    \label{it17}
\sum_i^K \exp{\{\mathbf{z}^T\mathbf{w}_i/\tau_T\}} \ge  
k \exp 0 = K
\end{equation}
Considering that the concavity of the exponential function in the region of interest (i.e. around zero) is relatively shallow, this approximation should be relatively good.

It should be noted that the cluster weight vectors do not have to be explicitly centred. By normalising them to unit vectors, their average is likely to be close to the origin regardless. 

For the normalising term in the denominator of (\ref{it15}) we can argue that for every $\mathbf{z}$, it can also be approximated by $K$. Hence the effect of the normalisation by cluster mixture in the numerator and the denominator will cancel out. The averaging over the distribution of embeddings in the denominator can again be simplified using Jensen's inequality. We can write
\begin{equation}
    \label{it18}
    \frac{1}{N}\sum_{j=1}^N \exp{\{\mathbf{z}_j^T\mathbf{w}_y/\tau_T \}}  \ge \exp{\{\bar{ \mathbf{z}}^T\mathbf{w}_y/\tau_T \}}
\end{equation}
where $\bar{\mathbf{z}}$ is the mean of the embedded data. 
Taking all these partial results into account, we can write
\begin{equation}
 \label{it18a}
Q(y|\mathbf{z}) \propto \exp{\{(\mathbf{z}- \bar{\mathbf{z}})^T\mathbf{w}_y/\tau_T\}}
\end{equation}
As the supervisory signal $Q(y|\mathbf{z})$ must satisfy the axiomatic properties of probabilities, the normalisation of the term on the right hand side of (\ref{it18a}) leads to  
\begin{equation}
\label{it19}
Q(y|\mathbf{z}) =\frac{\exp{\{(\mathbf{z}- \bar{\mathbf{z}})^T\mathbf{w}_y/\tau_T\}}}{\sum_{i=1}^K 
\exp{\{(\mathbf{z}- \bar{\mathbf{z}})^T\mathbf{w}_i/\tau_T\}}}
\end{equation}
Thus we have shown that the normalisation by inverse prior can be approaximated by data centering \cite{caron2018deep}. Through our analysis we have provided a theoretical justification for the centering operation, and in fact for the standard normalisation procedure by centering, used by most SSL methods.


A few comments are in order. The Jensen inequality provides a lower bound which can be quite loose for large magnitudes of the argument of the exponential function (logits). This would happen when the variation of the norm of embeddings and/or weight vectors is large. However, the data normalisation discussed in the previous section ensures that the norms are confined to a narrow interval.  The situation is likely to be aggravated by the choice of high temperature. However, if the inaccuracy of approximation is consistent for all training samples, as would be in the case of high temperature, then the probability distribution over cluster hypotheses will approach a one hot vector, and the approximation will not be invalidated. 


When the approximation by the lower bound in (\ref{it17}) does not quite hold, the term $\sum_i^K \exp{\{\mathbf{z}^T\mathbf{w}_i/\tau\}}$  cannot be cancelled out from the numerator and the denominator of (\ref{it15}). Instead, it acts as an inverse weight of the respective contributions of training samples to the value of the summation. The quality of the approximation will be particularly low for unambiguously clustered samples. As we replace $\sum_i^K \exp{\{\mathbf{z}^T\mathbf{w}_i/\tau\}}$ by the lower bound, which is significantly low, the weight of such strong samples will be magnified, with increased influence on the value of the integral. the resulting effect will be that this will further dampen down the teacher signal for strong clusters in favour of the weaker ones.

It is important to note, that, from the practical point of view, the above limitations are immaterial. In scenarios, where the approximations do not hold, the implication simply is that the relationship of normalisation by cluster priors to centering may break down. This does not impair centering's demonstrated effectiveness per se. However, it could explain its marginal inferiority to normalisation by cluster priors demnstrated experimentally by \cite{Amrani2022}.

Recalling (\ref{it8}) and (\ref{it19}) it is apparent that we can force the system to learn the supervisory signal in different ways. In all cases, it will involve some form of normalisation (mean, and variance and $L_2$ norm), and we will have several variants, depending whether the supervisory signal update is batch-based, running statistical parameter based or full training set-based. 

The first option is to scale the cluster posterior probabilities by their cluster conditional average as derived in (\ref{it8}). This mechanism involves a division by numbers whose values can be zero and will have to be regularised. As the supervisory signal takes values from $[0,1]$ and must satisfy $\sum_y Q(y|\mathbf{z})=1$, the rescaled posteriors have to be normalised. However, the update process itself is likely to be very noisy. 

The second option is to centralise the embeddings $\mathbf{z}$, as prescribed by (\ref{it19}). This can be done by a batch normalisation of the features applied  by the teacher before producing logits for the output probability computation. It is crucially important that the student network does not normalise the feature level embeddings, as this would prevent a proper learning, in spite of the temperature differences between the two networks. If both networks are batch normalised, the student and teacher tend to just copy each other's outputs. What we want the student network to do is to learn network parameters that will centralise the embedded data directly. It should be remembered that one of the assumptions behind this second option is that the weight vectors $\mathbf{w}_y$ are also centralised, i.e. $\mathbf{w}=0$. This can be achieved easily if we train the system with fixed weight vectors. However, even when the cluster models are updated, they are likely to average to zero, as they will span the centralised distribution of the embedded data. In any case, by virtue of the weight vectors being quasi orthogonal,  they will also tend to approach zero average.

It cannot be overemphasised that we are not promoting any particular normalisation method. In fact the flattening of the teacher and the predictor prior distributions can also be implemented using the minimum transport procedure. The actual choice of normalisation will depend on computational efficiency of the adopted procedure, its suitability for batch updating, and its expected performance. Our aim is simply to provide a theoretical underpinning for the process of  self-supervised learning by clustering.

In summary, Dino optimies regularised cross entropy, rather than our regularised K-L divergence. This does not impact on the learning of parameters $\theta$ and $\mathbf{w}_j$. However, their formulation does not lead to a closed form solution for the teacher normalisation by cluster priors, which in our case provides a bridge to the popular and successful normalisation by centering. \cite{Amrani2022} also used normalisation by cluster prior, but their argument is based on heuristics, rather than first principles. Interestingly, they confirmed by extensive  experimentation that normalisation by cluster priors does not lead to mode collapse. 

The closest to our formulation is the \cite{Sansone2025} lower bound on log joint probability distribution of the training samples and their augmented views, measured by K-L relative information loss, but they have not developed it further. Our formulation, starting from K-L divergence, and the derived closed form solution for the teacher distribution update,  provides an important insight into the effectioveness of the normalisation by centering. 

\section{Conclusions}
AI is developing extremely fast by means of experimental exploration, which focuses on novel deep neural network architectures, learning principles and strategies. The progress is driven by the relentless widening of the spectrum of applications and tasks considered, and by benchmarking, which rewards monotonic improvements in the performance achieved in different task categories. This quest for the development and discovery of exciting AI competences leaves the theoretical underpinning of the state of the art somewhat behind. This paper attempts to make a correction to this lopsided advancement of AI technology by presenting a theoretical model to underpin a specific family of popular and  effective self-supervised learning methods based on clustering, which mimic the popular solutions to unsupervised learning, i.e. the K-means algorithm and Gaussian mixture fitting, in particular. 

We formulate the problem of SSL as one of minimising the K-L divergence in a distillation learning setting, where both the student distribution, as well as the teacher are iteratively updated. This leads to optimising not only cross entropy of the teacher and student distributions, but entropy of the teacher distribution as well. The alternating direction optimisation regularised so as to avoid mode collapse leads to a distribution normalisation method in the form of scaling by inverse cluster probability prior. Most interestingly, using the approximations afforded by Jensen inequality, this normalisation method can be simplified to the popular normalisation method known as centering. 

The theoretical model developed is pleasing in a sense that it underpins the existing successful SSL methodology. It is also of pedagogic value for training new generations of researchers, because it links theory and practice. In addition, as any theory in general, it suggests a host of potential research directions to investigate in future. These include the idea of centering cluster weight vectors, learning with fixed weight vectors, the relationship of the dimensionality of deep representation and temperature, the choice of K, and effective ways of avoiding concept fragmentation, as well as cluster merging.      

{
    \small
    \bibliographystyle{ieeenat_fullname}
    \bibliography{main}

@String(ECCV= {Eur. Conf. Comput. Vis.})

@String(ICLR = {Int. Conf. Learn. Represent.})

@String(ECCV  = {ECCV})

@String(ICLR  = {ICLR})

@inproceedings{zhang2016colorful,
  title={Colorful image colorization},
  author={Zhang, Richard and Isola, Phillip and Efros, Alexei A},
  booktitle={European conference on computer vision},
  pages={649--666},
  year={2016},
  organization={Springer}
}

@article{Sansone2025,
  author       = {Emanuele Sansone and
                  Robin Manhaeve},
  title        = {Unifying Self-Supervised Clustering and Energy-Based Models},
  journal      = {Trans. Mach. Learn. Res.},
  year         = {2025}
}

@inproceedings{Amrani2022,
  author       = {Elad Amrani and
                  Leonid Karlinsky and
                  Alexander M. Bronstein},
  editor       = {Shai Avidan and
                  Gabriel J. Brostow and
                  Moustapha Ciss{\'{e}} and
                  Giovanni Maria Farinella and
                  Tal Hassner},
  title        = {Self-Supervised Classification Network},
  booktitle    = {Computer Vision - {ECCV} 2022 - 17th European Conference, Tel Aviv,
                  Israel, October 23-27, 2022, Proceedings, Part {XXXI}},
  series       = {Lecture Notes in Computer Science},
  volume       = {13691},
  pages        = {116--132},
  publisher    = {Springer},
  year         = {2022}
}

@article{Li2025,
  author       = {Jin Li and
                  Yaoming Wang and
                  Xiaopeng Zhang and
                  Dongsheng Jiang and
                  Wenrui Dai and
                  Chenglin Li and
                  Hongkai Xiong and
                  Qi Tian},
  title        = {Contrastive Learning via Variational Information Bottleneck},
  journal      = {{IEEE} Trans. Pattern Anal. Mach. Intell.},
  volume       = {47},
  number       = {9},
  pages        = {7410--7427},
  year         = {2025}
}

@inproceedings{Chen2020,
  author       = {Ting Chen and
                  Simon Kornblith and
                  Mohammad Norouzi and
                  Geoffrey E. Hinton},
  title        = {A Simple Framework for Contrastive Learning of Visual Representations},
  booktitle    = {Proceedings of the 37th International Conference on Machine Learning,
                  {ICML} 2020, 13-18 July 2020, Virtual Event},
  series       = {Proceedings of Machine Learning Research},
  volume       = {119},
  pages        = {1597--1607},
  publisher    = {{PMLR}},
  year         = {2020}
}

@inproceedings{Tan2024,
  author       = {Zhiquan Tan and
                  Jingqin Yang and
                  Weiran Huang and
                  Yang Yuan and
                  Yifan Zhang},
  title        = {Information Flow in Self-Supervised Learning},
  booktitle    = {Forty-first International Conference on Machine Learning, {ICML} 2024,
                  Vienna, Austria, July 21-27, 2024},
  publisher    = {OpenReview.net},
  year         = {2024}
}

@inproceedings{larsson2016learning,
  title={Learning representations for automatic colorization},
  author={Larsson, Gustav and Maire, Michael and Shakhnarovich, Gregory},
  booktitle={European conference on computer vision},
  pages={577--593},
  year={2016},
  organization={Springer}
}

@inproceedings{larsson2017colorization,
  title={Colorization as a proxy task for visual understanding},
  author={Larsson, Gustav and Maire, Michael and Shakhnarovich, Gregory},
  booktitle={Proceedings of the IEEE Conference on Computer Vision and Pattern Recognition},
  pages={6874--6883},
  year={2017}
}

@inproceedings{doersch2015unsupervised,
  title={Unsupervised visual representation learning by context prediction},
  author={Doersch, Carl and Gupta, Abhinav and Efros, Alexei A},
  booktitle={Proceedings of the IEEE international conference on computer vision},
  pages={1422--1430},
  year={2015}
}

@inproceedings{noroozi2016unsupervised,
  title={Unsupervised learning of visual representations by solving jigsaw puzzles},
  author={Noroozi, Mehdi and Favaro, Paolo},
  booktitle={European conference on computer vision},
  pages={69--84},
  year={2016},
  organization={Springer}
}

@inproceedings{kim2018learning,
  title={Learning image representations by completing damaged jigsaw puzzles},
  author={Kim, Dahun and Cho, Donghyeon and Yoo, Donggeun and Kweon, In So},
  booktitle={2018 IEEE Winter Conference on Applications of Computer Vision (WACV)},
  pages={793--802},
  year={2018},
  organization={IEEE}
}

@inproceedings{zhang2017split,
  title={Split-brain autoencoders: Unsupervised learning by cross-channel prediction},
  author={Zhang, Richard and Isola, Phillip and Efros, Alexei A},
  booktitle={Proceedings of the IEEE Conference on Computer Vision and Pattern Recognition},
  pages={1058--1067},
  year={2017}
}

@inproceedings{bojanowski2017unsupervised,
  title={Unsupervised learning by predicting noise},
  author={Bojanowski, Piotr and Joulin, Armand},
  booktitle={International Conference on Machine Learning},
  pages={517--526},
  year={2017},
  organization={PMLR}
}

@article{gidaris2018unsupervised,
  title={Unsupervised representation learning by predicting image rotations},
  author={Gidaris, Spyros and Singh, Praveer and Komodakis, Nikos},
  journal={arXiv preprint arXiv:1803.07728},
  year={2018}
}

@inproceedings{jenni2018self,
  title={Self-supervised feature learning by learning to spot artifacts},
  author={Jenni, Simon and Favaro, Paolo},
  booktitle={Proceedings of the IEEE Conference on Computer Vision and Pattern Recognition},
  pages={2733--2742},
  year={2018}
}

@article{simeoni2025dinov3,
  title={DINOv3},
  author={Sim{\'e}oni, Oriane and Vo, Huy V and Seitzer, Maximilian and Baldassarre, Federico and Oquab, Maxime and Jose, Cijo and Khalidov, Vasil and Szafraniec, Marc and Yi, Seungeun and Ramamonjisoa, Micha{\"e}l and others},
  journal={arXiv preprint arXiv:2508.10104},
  year={2025}
}

@inproceedings{caron2018deep,
  title={Deep clustering for unsupervised learning of visual features},
  author={Caron, Mathilde and Bojanowski, Piotr and Joulin, Armand and Douze, Matthijs},
  booktitle={Proceedings of the European conference on computer vision (ECCV)},
  pages={132--149},
  year={2018}
}

@inproceedings{chen2021exploring,
  title={Exploring simple siamese representation learning},
  author={Chen, Xinlei and He, Kaiming},
  booktitle={Proceedings of the IEEE/CVF conference on computer vision and pattern recognition},
  pages={15750--15758},
  year={2021}
}

@article{atito2021sit,
  title={{SiT}: Self-supervised vision transformer},
  author={Atito, Sara and Awais, Muhammad and Kittler, Josef},
  journal={arXiv preprint arXiv:2104.03602},
  year={2021}
}

@inproceedings{he2022masked,
  title={Masked autoencoders are scalable vision learners},
  author={He, Kaiming and Chen, Xinlei and Xie, Saining and Li, Yanghao and Doll{\'a}r, Piotr and Girshick, Ross},
  booktitle={Proceedings of the IEEE/CVF conference on computer vision and pattern recognition},
  pages={16000--16009},
  year={2022}
}

@inproceedings{xie2022simmim,
  title={{SimMIM}: A simple framework for masked image modeling},
  author={Xie, Zhenda and Zhang, Zheng and Cao, Yue and Lin, Yutong and Bao, Jianmin and Yao, Zhuliang and Dai, Qi and Hu, Han},
  booktitle={Proceedings of the IEEE/CVF conference on computer vision and pattern recognition},
  pages={9653--9663},
  year={2022}
}

@book{kullback1997information,
  title={Information theory and statistics},
  author={Kullback, Solomon},
  year={1997},
  publisher={Courier Corporation}
}

@article{grill2020bootstrap,
  title={Bootstrap your own latent-a new approach to self-supervised learning},
  author={Grill, Jean-Bastien and Strub, Florian and Altch{\'e}, Florent and Tallec, Corentin and Richemond, Pierre and Buchatskaya, Elena and Doersch, Carl and Avila Pires, Bernardo and Guo, Zhaohan and Gheshlaghi Azar, Mohammad and others},
  journal={Advances in neural information processing systems},
  volume={33},
  pages={21271--21284},
  year={2020}
}

@article{
oquab2023dinov2,
title={{DINO}v2: Learning Robust Visual Features without Supervision},
author={Maxime Oquab and Timoth{\'e}e Darcet and Th{\'e}o Moutakanni and Huy V. Vo and Marc Szafraniec and Vasil Khalidov and Pierre Fernandez and Daniel HAZIZA and Francisco Massa and Alaaeldin El-Nouby and Mido Assran and Nicolas Ballas and Wojciech Galuba and Russell Howes and Po-Yao Huang and Shang-Wen Li and Ishan Misra and Michael Rabbat and Vasu Sharma and Gabriel Synnaeve and Hu Xu and Herve Jegou and Julien Mairal and Patrick Labatut and Armand Joulin and Piotr Bojanowski},
journal={Transactions on Machine Learning Research},
issn={2835-8856},
year={2024},
url={https://openreview.net/forum?id=a68SUt6zFt},
note={Featured Certification}
}

@inproceedings{caron2021emerging,
  title={Emerging properties in self-supervised vision transformers},
  author={Caron, Mathilde and Touvron, Hugo and Misra, Ishan and J{\'e}gou, Herv{\'e} and Mairal, Julien and Bojanowski, Piotr and Joulin, Armand},
  booktitle={Proceedings of the IEEE/CVF international conference on computer vision},
  pages={9650--9660},
  year={2021}
}

@article{zhou2021ibot,
  title={iBOT: Image BERT Pre-Training with Online Tokenizer},
  author={Zhou, Jinghao and Wei, Chen and Wang, Huiyu and Shen, Wei and Xie, Cihang and Yuille, Alan and Kong, Tao},
  journal={International Conference on Learning Representations (ICLR)},
  year={2022}
}

@inproceedings{zbontar2021barlow,
  title={Barlow twins: Self-supervised learning via redundancy reduction},
  author={Zbontar, Jure and Jing, Li and Misra, Ishan and LeCun, Yann and Deny, St{\'e}phane},
  booktitle={International conference on machine learning},
  pages={12310--12320},
  year={2021},
  organization={PMLR}
}

@article{bardes2022vicregl,
  title={Vicregl: Self-supervised learning of local visual features},
  author={Bardes, Adrien and Ponce, Jean and LeCun, Yann},
  journal={Advances in Neural Information Processing Systems},
  volume={35},
  pages={8799--8810},
  year={2022}
}

@inproceedings{stegmuller2023croc,
  title={Croc: Cross-view online clustering for dense visual representation learning},
  author={Stegm{\"u}ller, Thomas and Lebailly, Tim and Bozorgtabar, Behzad and Tuytelaars, Tinne and Thiran, Jean-Philippe},
  booktitle={Proceedings of the IEEE/CVF Conference on Computer Vision and Pattern Recognition},
  pages={7000--7009},
  year={2023}
}

@inproceedings{
lebailly2024cribo,
title={Cr{IB}o: Self-Supervised Learning via Cross-Image Object-Level Bootstrapping},
author={Tim Lebailly and Thomas Stegm{\"u}ller and Behzad Bozorgtabar and Jean-Philippe Thiran and Tinne Tuytelaars},
booktitle={The Twelfth International Conference on Learning Representations},
year={2024},
}

@article{xie2021unsupervised,
  title={Unsupervised object-level representation learning from scene images},
  author={Xie, Jiahao and Zhan, Xiaohang and Liu, Ziwei and Ong, Yew Soon and Loy, Chen Change},
  journal={Advances in Neural Information Processing Systems},
  volume={34},
  pages={28864--28876},
  year={2021}
}

@inproceedings{henaff2021efficient,
  title={Efficient visual pretraining with contrastive detection},
  author={H{\'e}naff, Olivier J and Koppula, Skanda and Alayrac, Jean-Baptiste and Van den Oord, Aaron and Vinyals, Oriol and Carreira, Joao},
  booktitle={Proceedings of the IEEE/CVF International Conference on Computer Vision},
  pages={10086--10096},
  year={2021}
}

@inproceedings{
ukic2025objectcentricOCEBO,
title={Object-Centric Pretraining via Target Encoder Bootstrapping},
author={Nikola Duki{\'c} and Tim Lebailly and Tinne Tuytelaars},
booktitle={The Thirteenth International Conference on Learning Representations},
year={2025},
}

@inproceedings{wu2018unsupervised,
  title={Unsupervised feature learning via non-parametric instance discrimination},
  author={Wu, Zhirong and Xiong, Yuanjun and Yu, Stella X and Lin, Dahua},
  booktitle={Proceedings of the IEEE conference on computer vision and pattern recognition},
  pages={3733--3742},
  year={2018}
}

@inproceedings{ioffe2015batch,
  title={Batch normalization: Accelerating deep network training by reducing internal covariate shift},
  author={Ioffe, Sergey and Szegedy, Christian},
  booktitle={International conference on machine learning},
  pages={448--456},
  year={2015},
  organization={pmlr}
}

@article{caron2020swav,
  title={Unsupervised Learning of Visual Features by Contrasting Cluster Assignments},
  author={Caron, Mathilde and Misra, Ishan and Mairal, Julien and Goyal, Priya and Bojanowski, Piotr and Joulin, Armand},
  journal={NeurIPS},
  year={2020}
}

@article{Jain-prl2010,
  author       = {Anil K. Jain},
  title        = {Data clustering: 50 years beyond K-means},
  journal      = {Pattern Recognit. Lett.},
  volume       = {31},
  number       = {8},
  pages        = {651--666},
  year         = {2010}
}

@inproceedings{MacQueen,
author={MacQueen, J},
year={1967},
title={Some methods for classification and analysis of multivariate
observations},
booktitle={Fifth Berkeley Symposium on Mathematics. Statistics and
Probability},
publisher={University of California Press},  pages={281–297}
}

@article{Forgy,
author={Forgy, EW},
title={Cluster analysis of multivariate data: Efficiency vs.
interpretability of classifications}, journal={Biometrics},
volume={21},
pages={768–769},
year={1965}
}

@article{Rubin,
author={Dempster, AP and Laird, NM and Rubin, DB},
title={Maximum likelihood from incomplete
data via the EM algorithm},
year={1977},
journal={Journal Roy. Statist. Soc. Series B},
volume={39},
pages={1–38}
}

@inproceedings{Asano-iclr2020,
  author       = {Yuki Markus Asano and
                  Christian Rupprecht and
                  Andrea Vedaldi},
  title        = {Self-labelling via simultaneous clustering and representation learning},
  booktitle    = {8th International Conference on Learning Representations, {ICLR} 2020,
                  Addis Ababa, Ethiopia, April 26-30, 2020},
  publisher    = {OpenReview.net},
  year         = {2020}
}

@article{Pearson,
title={Contributions to the mathematical theory of evolution},
author={Pearson, Karl},
journal={ Phil. Trans. Roy. Soc. London A },
volume={185},
pages={71-110},
year={1894}
}

@article{fukushima1980neocognitron,
  title={Neocognitron: A self-organizing neural network model for a mechanism of pattern recognition unaffected by shift in position},
  author={Fukushima, Kunihiko},
  journal={Biological Cybernetics},
  volume={36},
  number={4},
  pages={193--202},
  year={1980}
}

@article{wiskott2002sfa,
  title={Slow feature analysis: Unsupervised learning of invariances},
  author={Wiskott, Laurenz and Sejnowski, Terrence J},
  journal={Neural Computation},
  volume={14},
  number={4},
  year={2002}
}

@article{hinton2006dbn,
  title={A fast learning algorithm for deep belief nets},
  author={Hinton, Geoffrey E and Osindero, Simon and Teh, Yee-Whye},
  journal={Neural Computation},
  volume={18},
  number={7},
  pages={1527--1554},
  year={2006}
}

@article{oord2018cpc,
  title={Representation learning with contrastive predictive coding},
  author={van den Oord, A{\"a}ron and Li, Yazhe and Vinyals, Oriol},
  journal={arXiv preprint arXiv:1807.03748},
  year={2018}
}

@inproceedings{saunshi2019theory,
  title={A theoretical analysis of contrastive unsupervised representation learning},
  author={Saunshi, Nikunj and Plevrakis, Orestis and Arora, Sanjeev and Khodak, Mikhail and Khandeparkar, Hrishikesh},
  booktitle={International Conference on Machine Learning},
  year={2019}
}

@inproceedings{Tishby1999,
    author={Tishby, Naftali and Pereira, Fernando and Bialek, William} ,
    title={The information bottleneck method},
    booktitle={The 37th Annual Conf. on Communication, Control and Computing},
pages={368-377},
    year={1999} 
}

@article{Vincent2010,
author={Vincent, Pascal and Larochelle, Hugo},
title={Stacked Denoising Autoencoders: Learning Useful Representations in a Deep Network with a Local Denoising Criterion},
journal={Journal of Machine Learning Research},
volume={11},
pages={3371-3408},
year={2010}
}
}




\end{document}